\documentclass[letterpaper]{article} 
\usepackage{aaai25}  
\usepackage{times}  
\usepackage{helvet}  
\usepackage{courier}  
\usepackage[hyphens]{url}  
\usepackage{graphicx} 
\usepackage{natbib}  
\usepackage{caption} 
\usepackage{subcaption}
\usepackage{multirow}
\usepackage{booktabs}
\usepackage{array}
\usepackage{algorithm}
\usepackage{algorithmic}
\usepackage{amsmath}
\usepackage{amssymb}
\usepackage{mathtools}
\usepackage{tcolorbox}
\usepackage{graphicx}
\graphicspath{{LaTeX/images/}{images/}}

\usepackage{newfloat}
\usepackage{listings}
\DeclareCaptionStyle{ruled}{labelfont=normalfont,labelsep=colon,strut=off} 
\floatstyle{ruled}
\newfloat{listing}{tb}{lst}{}
\floatname{listing}{Listing}
\nocopyright
\title{Mapping from Meaning: Addressing the Miscalibration of Prompt-Sensitive Language Models}
\author{
    Kyle Cox\textsuperscript{\rm 1},
    Jiawei Xu\textsuperscript{\rm 1},
    Yikun Han\textsuperscript{\rm 2},
    Rong Xu\textsuperscript{\rm 1},
    Tianhao Li\textsuperscript{\rm 1},
    Chi-Yang Hsu\textsuperscript{\rm 1},
    Tianlong Chen\textsuperscript{\rm 3},
    Walter Gerych\textsuperscript{\rm 4},
    Ying Ding\textsuperscript{\rm 1}
}
\affiliations{
    \textsuperscript{\rm 1}University of Texas at Austin\\
    \textsuperscript{\rm 2}University of Michigan-Ann Arbor\\
    \textsuperscript{\rm 3}University of North Carolina at Chapel Hill\\
    \textsuperscript{\rm 4}MIT

    \{kylecox, jiaweixu, rongxu, tianhao, ch52669\}@utexas.edu,
    yikunhan@umich.edu,
    tianlong@cs.unc.edu,
    wgerych@mit.edu,
    ying.ding@ischool.utexas.edu
}

\begin{document}
\frenchspacing
\maketitle
\begin{abstract}
An interesting behavior in large language models (LLMs) is prompt sensitivity. When provided with different but semantically equivalent versions of the same prompt, models may produce very different distributions of answers. This suggests that the uncertainty reflected in a model's output distribution for one prompt may not reflect the model's uncertainty about the \textit{meaning} of the prompt. We model prompt sensitivity as a type of generalization error, and show that sampling across the semantic ``concept space'' with paraphrasing perturbations improves uncertainty calibration without compromising accuracy. Additionally, we introduce a new metric for uncertainty decomposition in black-box LLMs that improves upon entropy-based decomposition by modeling semantic continuities in natural language generation. We show that this decomposition metric can be used to quantify how much LLM uncertainty is attributed to prompt sensitivity. Our work introduces a new way to improve uncertainty calibration in prompt-sensitive language models, and provides evidence that some LLMs fail to exhibit consistent general reasoning about the meanings of their inputs.
\end{abstract}

\begin{links}
\link{Code}{https://github.com/xocelyk/paraphrase-uncertainty}
\end{links}

\begin{figure*}
    \centering
    \includegraphics[width=1\linewidth]{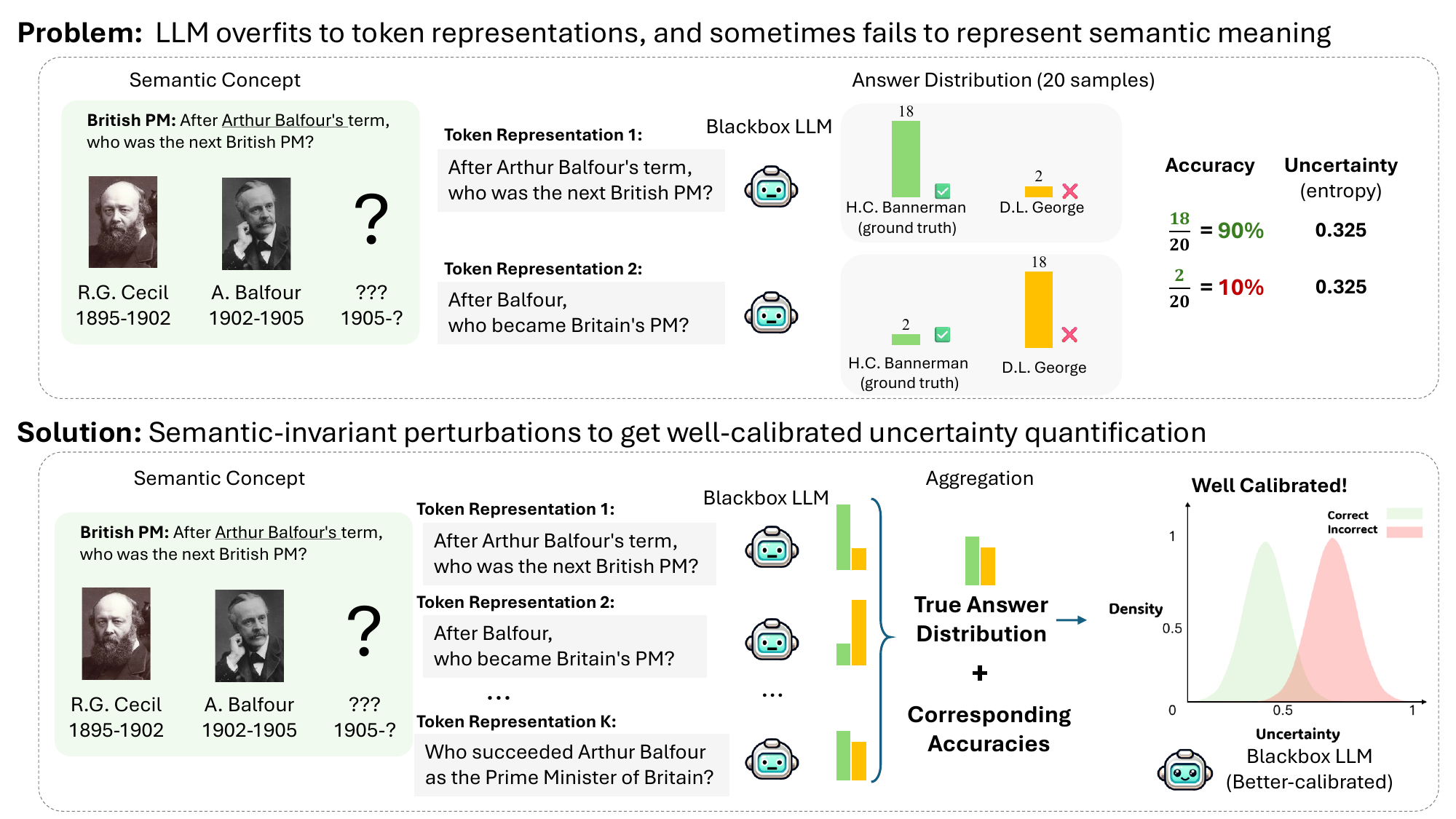}
    \caption{Problem statement and the semantic-invariant perturbation framework.}
    \label{fig:1}
\end{figure*}
\section{Introduction}
As the size of large language models (LLMs) has scaled, so has their performance across benchmark tasks~\cite{openai2024gpt4, kaplan2020scalinglawsneurallanguage}. Such gains have led many to question whether LLMs are exhibiting early signs of general intelligence \cite{bubeck2023sparks}. However, recent work has shown that the distinction between reasoning and memorization is nuanced. Many of the behaviors that appear to indicate general reasoning ability in LLMs in fact represent modes of memorization. For instance, GPT-4's ability to decode ROT13 ciphers but not less common variants like ROT2 raises questions about how LLMs acquire and generalize skills  \cite{mccoy2023embersautoregressionunderstandinglarge}. Similar patterns emerge in other capabilities. \citet{lu2024emergentabilitieslargelanguage} observe that some LLMs can solve arithmetic word problems and demonstrate physical intuition through in-context learning, but struggle in zero-shot settings, suggesting a dependence on pattern recognition. As new capabilities emerge, the nature of each advancement—whether it represents true reasoning ability or increasingly sophisticated forms of memorization—remains a fundamental question in AI research.

A relevant phenomenon in large language models is prompt sensitivity \cite{Sclar2023Sensitivity, Lu2022Fantastically, chen2024premise, shi2023large}. Simply reformatting prompts by turning uppercase characters to lowercase or modifying spaces between input fields can result in dramatic differences in responses~\cite{Sclar2023Sensitivity}. Further, changing the order of examples in few-shot learning can lead to substantial variance in accuracy, even in very large models ~\cite{Lu2022Fantastically}. We consider whether prompt sensitivity in LLMs also indicates a failure of general reasoning: Is the model's response a function of the \textit{meaning} of a prompt, or merely the sequence of tokens that convey it?

We address this question with an uncertainty quantification framework, analyzing how distributional properties of LLM outputs vary under meaning-preserving perturbations. While models exhibiting true general reasoning should maintain consistent distributions across semantically equivalent inputs, memorization-based models may produce distributions that reflect surface-level statistical patterns in their training data rather than deeper semantic understanding. This distinction results in differences in calibration, even when accuracy is held constant across both types of models. We leverage this insight to develop a calibration-based measure of general reasoning performance. Uncertainty calibration offers a unique way to assess a model's general reasoning performance on in-domain tasks.

To this end, we develop a framework of semantic-invariant perturbations, transforming input questions while preserving their meaning. We introduce a new uncertainty quantification metric that captures semantic continuities in language and can be decomposed into epistemic uncertainty (reflecting uncertainty about meaning) and aleatoric uncertainty (reflecting uncertainty about token sequences). We show that sampling with our perturbation framework and measuring uncertainty with our total variance metric yields superior calibration on question-answering (QA) tasks, supporting the hypothesis that some LLMs overfit to token sequences. We also show that decomposing this metric into epistemic uncertainty and aleatoric uncertainty can be used to quantify distributional sensitivity to meaning-preserving perturbations.

\section{Overfitting in Question Answering}

Figure \ref{fig:1} demonstrates the problem we address in this paper: overfitting in question answering. In question answering, it is important to distinguish between the semantic content and the lexical representation of a question. The answer to a question is fundamentally a function of its meaning rather than the specific language used to express it. We refer to this underlying meaning as the \textit{semantic concept} and the specific words as the \textit{token representation}. LLMs are trained to predict future tokens based on sequences of tokens, thus directly learning the map from token representations to answers. But models that generalize well should also learn the map from semantic concepts to answers.

Previous work has shown that neural networks can abruptly transition from memorization to discovering general patterns. For instance, transformers trained on modular arithmetic initially memorize examples before learning generalizable arithmetic rules \cite{power2022grokking}. In question answering, we propose evaluating generalization performance by measuring how a model's answer distribution varies across different token representations of the same semantic concept. A model that has learned true semantic patterns should maintain consistent answer distributions regardless of phrasing. However, when models generate varying distributions for semantically equivalent questions, this indicates potential overfitting to specific token sequences. This token-level overfitting impairs the model's ability to reliably represent uncertainty, resulting in poor uncertainty calibration.

Prior work has shown that foundation LLMs do indeed overfit, responding in different ways to semantically equivalent prompts \cite{kadavath2022language, Sclar2023Sensitivity, chen2024premise, shi2023large}. Given this problem, we ask two questions: (1) \textit{How can we recover uncertainty calibration from an overfitted model?} and (2) \textit{How can we quantify how much a model overfits?}

\section{Semantic-Invariant Perturbation Sampling}

\textbf{Perturbation framework. }Given that a model overfits to unique token representations, we aim to recover the model's answer distribution over the entire semantic concept. In the context of a black-box LLM, where we can only measure distributions by taking samples, we propose sampling across the semantic concept space rather than from individual token representations.

\begin{figure*}[ht]
    \centering
    \includegraphics[width=1\linewidth]{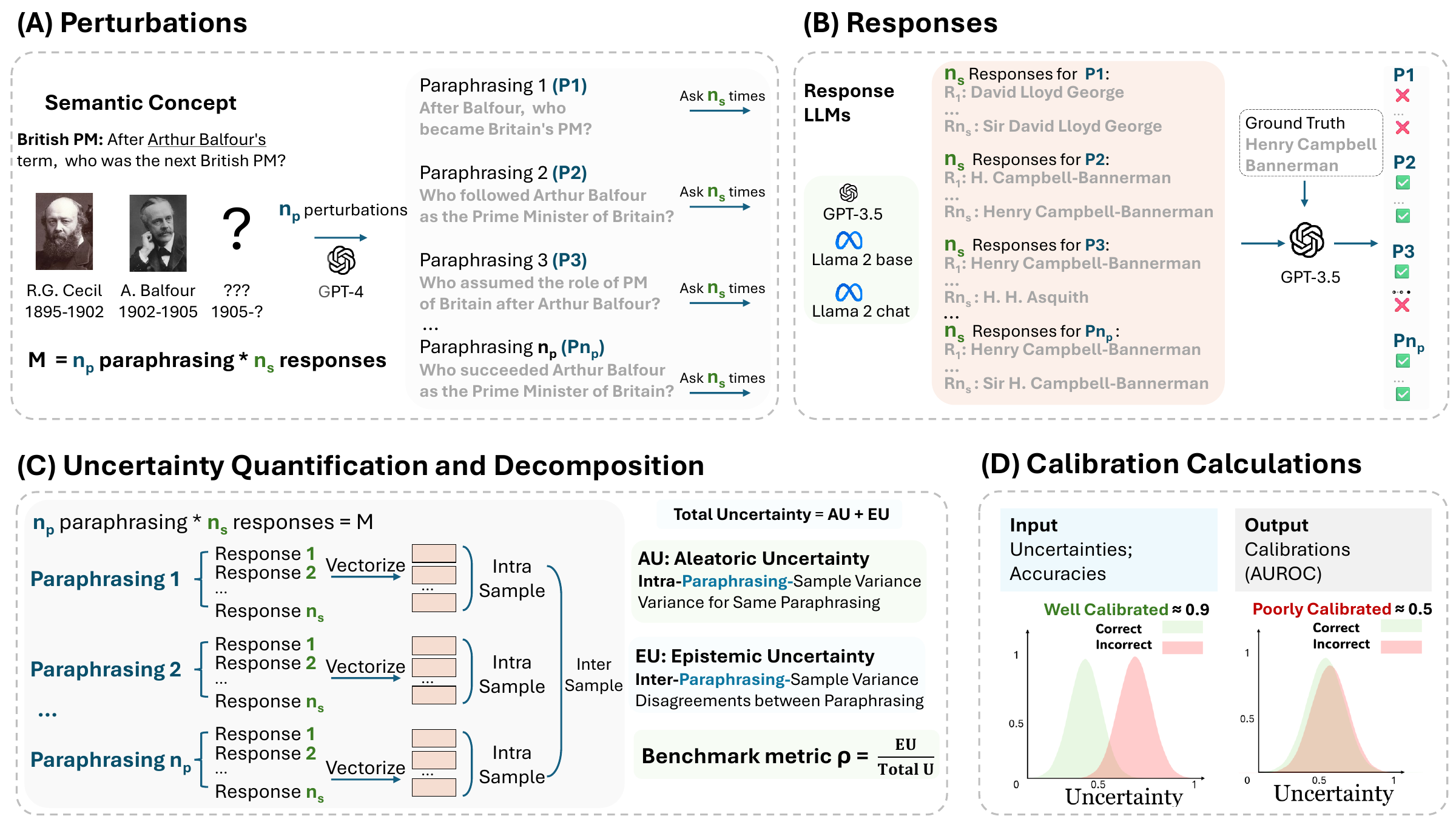}
    \caption{Pipeline for the semantic-invariant perturbation framework.}
    \label{fig:2}
\end{figure*}

To achieve this, we use semantic-invariant perturbations, specifically through paraphrasing. Paraphrasing involves generating multiple different but semantically equivalent versions of the same question. By sampling the model's responses to these varied paraphrases, we can estimate a more accurate answer distribution that reflects the underlying semantic concept rather than the specific wording. This method mitigates the overfitting issue by averaging the model's outputs across different phrasings, thus better measuring its true uncertainty and improving calibration. Figure \ref{fig:2} illustrates the semantic-invariant perturbation framework.

For example, if the original question were ``Who was the British Prime Minister after Arthur Balfour?", we generate $n_{\text{p}}$ paraphrases such as ``Who succeeded Arthur Balfour as Prime Minister of Britain?" and ``After Arthur Balfour, who became the British Prime Minister?" We then take $n_{\text{s}}$ samples of the model's response to each of these paraphrases and aggregate the results to form a comprehensive distribution over the semantic concept.

\textbf{Theoretical motivation.} We borrow from ergodic theory to illustrate our method. Suppose that the semantic concept space is the union of its $m$ token representations. We assume that each token representation is equally likely, i.e., that token representations take up equal space in the semantic concept space. It is useful to model the semantic concept space as the circle $S^1$ and each token representation $t_i$ as a connected interval in $S^1$ from $\left[\frac{i-1}{m}, \frac{i}{m}\right)$.

Each token representation is modeled by a random variable $Q(t_i)$ representing the language model's distribution over responses at that token representation. Under the assumption that each token representation is equally likely, we claim that $Q(c)$, the model's true response distribution over the semantic concept, is equal to the average distribution at each token representation:
\begin{equation}
\fontsize{7pt}{8pt}\selectfont
Q(c) = \frac{1}{m} \sum_{i=1}^m Q(t_i).
\label{ergodic-1}
\end{equation}
We say that Equation \ref{ergodic-1} represents the \textit{space average} over the semantic concept space.

We consider randomly sampling token representations from the semantic concept space as a map from a previous token representation to the next. We define this map by an irrational circle rotation:
\begin{equation}
\fontsize{7pt}{8pt}\selectfont
T_\theta: [0, 1] \to [0, 1], \quad T_\theta(x) = (x + \theta) \mod 1, \quad \theta \in \mathbb{R} \setminus \mathbb{Q}.
\label{ergodic-2}
\end{equation}
This map is ergodic \cite{Hawkins2021-ak}. Over infinitely many samples, the map will visit each token representation with a frequency proportional to its size in the concept space. By construction, the representations are of equal size, so this sampling map limits to the uniform distribution over token representations.

By the Birkhoff ergodic theorem, we notice that the time average along the trajectory defined by $T_\theta$ limits to the space average of the semantic concept space as time goes to infinity \cite{Hawkins2021-ak}:
\begin{equation}
\fontsize{7pt}{8pt}\selectfont
\label{ergodic-3}
\lim_{n \to \infty} \frac{1}{n}\sum_{k=0}^{n-1}Q(T^k(x)) = \frac{1}{m} \sum_{i=1}^m Q(t_i) = Q(c).
\end{equation}
In our perturbation method, this means that randomly uniformly sampling over the token representations, measuring the response distribution $Q(t_i)$ at each representation, and averaging over time approaches the true distribution we aim to model, $Q(c).$ In practice, we assume a black-box model, so we cannot exactly measure $Q(t_i)$ at each step. Instead, we sample some small number of times from $Q(t_i)$ at each step, and the rate of convergence depends upon the sampling efficiency at each $t_i.$

\section{Uncertainty Decomposition}
Introducing perturbations for each question and sampling from those perturbations can be interpreted as decomposing the total uncertainty into components of epistemic uncertainty and aleatoric uncertainty. As illustrated in Figure \ref{fig:2} (C), epistemic uncertainty captures the inter-sample uncertainty, i.e., the uncertainty introduced by including different perturbations, and aleatoric captures the intra-sample uncertainty, i.e., the uncertainty at a particular token representation. Under our perturbation framework, epistemic uncertainty can be interpreted as \textit{paraphrase uncertainty}, or the uncertainty attributed to the model's prompt sensitivity, and aleatoric uncertainty can be interpreted as \textit{sampling uncertainty}, or the average uncertainty reflected in the model's output distribution at specific prompts.

Quantifying each of these components yields important insights. In a well-calibrated model, epistemic uncertainty should make up a small fraction of its total uncertainty, because the distributions it generates at different paraphrases should be similar. The ratio of epistemic uncertainty to total uncertainty quantifies how sensitive the model is to its input phrasings. Further, by calculating the calibration attributed to epistemic and aleatoric uncertainty independently, we can measure the marginal calibration improvement due to adding a new paraphrase to the sampling pipeline. To quantify each of these components, we decompose total uncertainty into its constituent parts.

A desirable property of an uncertainty metric \( U \) is that the metric decomposes additively by: 
\begin{equation}
\fontsize{7pt}{8pt}\selectfont
U = U_{\text{e}} + U_{\text{a}}
\label{uq-decomp}
\end{equation}
such that $U_{\text{e}}$ represents the inter-sample (epistemic) uncertainty and $U_{\text{a}}$ represents the intra-sample (aleatoric) uncertainty. Previous work on uncertainty decomposition in LLMs has used entropy-based metrics to decompose uncertainty in this way \cite{kuhn2023semantic, hou2023decomposing}. However, entropy-based metrics require discretizing LLM outputs into distinct classes and calculating entropy over the resulting probability distribution. This discretization disregards semantic continuities and partial similarities, and consequently results in suboptimal calibration. We introduce a new uncertainty metric for black-box LLMs based on embedding variance, which models semantic continuities while decomposing additively as described by Equation \ref{uq-decomp}.

\subsection{Background on Uncertainty Quantification}
\label{uq}
Our framework builds upon work that quantifies uncertainties in \textit{meaning} for natural language generation (NLG). In NLG, quantifying uncertainty is crucial for interpreting and improving model performance. Calibration is a common way to evaluate an uncertainty measure.

\textbf{Calibration.} A useful uncertainty metric is one that provides information about whether the model's prediction is correct or incorrect. An uncertainty metric with high calibration will result in very different uncertainty distributions for correct generations and incorrect generations (see part (D) of Figure \ref{fig:2}). Following previous work \cite{kuhn2023semantic, lin2023generating}, we evaluate uncertainty metrics by using them to predict if a response was correct and calculating the Area Under Receiver Operating Characteristic (AUROC) of this function.

\textbf{Entropy.} Traditional uncertainty metrics, such as entropy, measure the unpredictability of a model's output distribution \cite{Lu2022Fantastically}. The uncertainty for a classifier given input $x$ can be measured by the entropy of its output distribution $Y$:
\begin{equation}
\fontsize{7pt}{8pt}\selectfont
U_{\text{entropy}}(x) = H(Y \mid x) = -\sum_{y \in Y} P(y \mid x) \ln \big( P(y \mid x) \big).
\label{entropy}
\end{equation}
\textbf{Semantic entropy.} However, in NLG, the model may generate lexically distinct but semantically equivalent responses, such as ``The United States" and ``USA." To address this, \citet{kuhn2023semantic} introduced semantic entropy, which calculates entropy over classes of semantically equivalent responses $c \in C$:
\begin{equation}
\fontsize{7pt}{8pt}\selectfont
U_{\text{SE}}(x) = H(Y \mid x) = -\sum_{c \in C} P(c \mid x) \ln \big(P(c \mid x)
\big).
\label{semantic-entropy}
\end{equation}
\textbf{Affinity graph metrics.} While semantic entropy improves upon traditional entropy by considering semantic equivalence, it still relies on a discrete model of semantic space. In practice, responses are often similar in meaning but not identical. For white-box settings, \citet{duan2024shiftingattentionrelevancepredictive} proposed an attention mechanism to measure how much each token contributes to meaning and compute a weighted average of uncertainty. To capture these continuous relationships in white-box models, \citet{lin2023generating} propose creating an affinity graph, where nodes represent unique responses and edges represent relationships (bidirectional entailment, contradiction, or similarity) between pairs of responses. One metric derived from the graph's Laplacian, the sum of eigenvalues, extends the concept of semantic sets to \textit{continuous} semantic sets (\( U_{\text{EigV}}\)). The affinity-graph metrics proposed by \citet{lin2023generating} improve upon previous metrics by modeling semantic continuities when quantifying uncertainty, and are generally better calibrated than entropy and semantic entropy (see Section \ref{calibration-results}).


\subsection{Background on Uncertainty Decomposition}
\label{decomposition-background}
Affinity-graph-based uncertainty metrics improve calibration in NLG, but by diverging from conventional entropy-based metrics they sacrifice entropy's interpretability. A valuable property of entropy-based uncertainty quantification metrics is the ability to decompose into aleatoric and epistemic uncertainty:
\begin{equation}
\fontsize{7pt}{8pt}\selectfont
H(Y \mid x) = \underbrace{H(Y \mid x, \theta)}_{\text{aleatoric}} + \underbrace{I(Y, \theta \mid x)}_{\text{epistemic}},
\label{entropy-decomposition}
\end{equation}
where \(\theta\) represents a particular parameterization. In Bayesian Neural Networks, this decomposition reflects that models are randomly parameterized and there is uncertainty in these parameters \cite{depeweg2018decompositionuncertaintybayesiandeep}. In the context of LLMs, we suppose $\theta$ represents that for any question $x,$ there is a distribution over possible ways to phrase that question.

This decomposition has been used for ambiguity detection, in-context learning analysis, and calibration evaluation \cite{hou2023decomposing, ling2024uncertainty}. However, entropy-based decomposition is not easily generalized to newer metrics. \citet{ling2024uncertainty} suggest decomposing uncertainty via the law of total variance:
\begin{equation}
\fontsize{7pt}{8pt}\selectfont
\text{Var}(Y \mid x) = \underbrace{\mathbb{E}_{\theta}\left[\text{Var}(Y \mid x, \theta)\right]}_{\text{aleatoric}} + \underbrace{\text{Var}\big(\mathbb{E}_{\theta}[Y \mid x, \theta]\big)}_{\text{epistemic}}.
\label{variance-decomposition}
\end{equation}
However, taking variance over a probability distribution of outputs does not resolve the issue of discretizing the semantic output space. Instead, we propose a generalization of variance decomposition to the many-dimensional case.

\subsection{Uncertainty Decomposition by Embedding Variance}
\label{embedding-variance}
Rather than force discrete classes on our outputs, we take each output and embed it, adopting the classic continuous, many-dimensional model of semantic space. We then calculate the covariance matrix of these embeddings and take its trace. Leveraging the notion of embedding dimensions as latent features, we interpret this metric as the total dispersion over our basis of latent features. Part (C) of Figure \ref{fig:2} visually represents uncertainty decomposition using the embedding variance method.

Let \(\textbf{Y} \in \mathbb{R}^{m \times d}\) be the matrix of embeddings, where \( m \) is the number of samples and \( d \) is the embedding dimension. The law of total covariance can be described as follows.

\begin{equation}
\fontsize{7pt}{8pt}\selectfont
\underbrace{\text{Cov} \left(\mathbf{Y} \mid x \right)}_{\text{total}} = \underbrace{\mathbb{E}_\theta \left[ \text{Cov}\left(\mathbf{Y} \mid x, \theta \right)\right]}_{\text{aleatoric}} + \underbrace{\text{Cov} \left( \mathbb{E}_\theta[\mathbf{Y} \mid x, \theta] \right)}_{\text{epistemic}}.
\label{covariance-decomposition}
\end{equation}

We can restate this equation in terms of the covariance matrices \( \mathbf{\Sigma}_{t}, \mathbf{\Sigma}_{a}, \) and \( \mathbf{\Sigma}_{e} \). However, we would like to express uncertainty and its components as real values. We note that each \( \sigma_{ii} \in \mathbf{\Sigma}_{t} \) represents the variance of the sample over the embedding dimension \( i \), and \( \text{tr}(\mathbf{\Sigma}_{t}) = \sum_i^d \sigma_{ii} \) represents the total variance over all embedding dimensions, which is an informative uncertainty metric.
\begin{equation}
\fontsize{7pt}{8pt}\selectfont
\begin{split}
\mathbf{\Sigma}_{t} &= \mathbf{\Sigma}_{a} + \mathbf{\Sigma}_{e} \\
\text{tr}(\mathbf{\Sigma}_{t}) &= \text{tr}(\mathbf{\Sigma}_{a}) + \text{tr}(\mathbf{\Sigma}_{e}) \\
U_{t} &= U_{a} + U_{e}
\end{split}
\label{covariance-metric}
\end{equation}
\(U_{t}\) is our decomposable embedding variance uncertainty metric, described as ``Variance (Total)" in Table~\ref{Table:Data1_overall}. The embedding variance uncertainty metric can be applied to any embedding space, and may be particularly useful for evaluating foundation models on domain-specific tasks with a specialized embedding model. Our evaluations involve more conventional question answering. For evaluation, we embed using the eigenvectors of the graph Laplacian following~\citet{lin2023generating} and~\citet{ng2001spectral}. 

\begin{table*}
	\centering
	\setlength{\tabcolsep}{.9mm} 
    \fontsize{9}{11}\selectfont
	\begin{tabular}{clccccccccc}
		\toprule
		&{\multirow{2}{*}{Metric}} & \multicolumn{4}{c}{m = 6} & \multicolumn{5}{c}{ m = 12}  \\
		\cmidrule(lr){3-6} \cmidrule(lr){7-11}
		\multicolumn{2}{c}{} & $(1,6)$ & $(2,3)$ & $(3,2)$ & $(6,1)$ & $(1,12)$ & $(2,6)$ & $(3,4)$ & $(4,3)$ & $(6,2)$ \\
		\midrule
		\multicolumn{11}{c}{TriviaQA}\\
		\midrule
		\multirow{8}{*}{\begin{tabular}[c]{@{}c@{}}GPT 3.5 \end{tabular}} 
        & Entropy & 74.3$\pm$0.8 & 77.7$\pm$1.7 & 77.8$\pm$0.6 & 78.2$\pm$1.0 & 77.4$\pm$0.8 & 78.8$\pm$0.9 & 79.5$\pm$0.5 & 80.1$\pm$1.1 & 79.2$\pm$1.0 \\
        & LexiSim & 77.3$\pm$0.6 & 81.0$\pm$1.8 & 81.6$\pm$0.7 & 82.0$\pm$1.1 & 80.3$\pm$1.0 & 82.1$\pm$1.0 & 83.4$\pm$0.6 & 83.6$\pm$1.0 & 83.1$\pm$1.1 \\
        & SE & 75.6$\pm$0.7 & 79.5$\pm$2.0 & 79.6$\pm$1.0 & 80.3$\pm$1.3 & 80.1$\pm$1.1 & 81.2$\pm$1.1 & 82.1$\pm$0.6 & 83.1$\pm$1.2 & 82.0$\pm$0.8 \\
        & $U_{\text{EigV}}$ & \textbf{79.2$\pm$0.8} & \textbf{83.3$\pm$2.0} & 84.2$\pm$0.8 & 84.7$\pm$1.1 & 83.3$\pm$1.2 & 84.7$\pm$0.9 & 85.7$\pm$0.7 & 86.5$\pm$1.1 & 85.6$\pm$0.7 \\
        & Variance (Tot) & 79.0$\pm$1.2 & 82.9$\pm$2.2 & \textbf{84.3$\pm$0.9} & \textbf{84.9$\pm$1.2} & \textbf{83.5$\pm$1.1} & \textbf{85.0$\pm$0.8} & \textbf{86.4$\pm$0.7} & \textbf{87.4$\pm$1.0} & \textbf{86.4$\pm$0.7} \\
        & Variance (AU) & 79.0$\pm$1.2 & 82.7$\pm$2.2 & 84.1$\pm$0.8 & --- & \textbf{83.5$\pm$1.1} & 84.9$\pm$0.8 & 86.3$\pm$0.7 & 86.9$\pm$0.9 & 85.7$\pm$0.8 \\
        & Variance (EU) & --- & 81.9$\pm$2.6 & 83.7$\pm$1.1 & \textbf{84.9$\pm$1.2} & --- & 83.4$\pm$1.2 & 85.8$\pm$0.5 & 87.0$\pm$0.7 & 86.2$\pm$0.8 \\
		\midrule
		\multicolumn{11}{c}{NQ}\\
		\midrule
		\multirow{8}{*}{\begin{tabular}[c]{@{}c@{}} GPT 3.5 \end{tabular}} 
        & Entropy & 54.6$\pm$1.1 & 54.0$\pm$0.9 & 54.0$\pm$0.7 & 54.3$\pm$1.0 & 54.4$\pm$0.3 & 54.4$\pm$0.3 & 54.7$\pm$1.0 & 54.1$\pm$0.5 & 54.5$\pm$0.4 \\
        & LexiSim & 62.3$\pm$1.3 & 62.6$\pm$0.9 & 64.0$\pm$1.0 & 64.5$\pm$1.1 & 63.1$\pm$1.2 & 63.6$\pm$0.7 & 64.5$\pm$1.0 & 64.4$\pm$0.4 & 65.1$\pm$0.2 \\
        & SE & 60.2$\pm$1.1 & 61.5$\pm$0.6 & 62.1$\pm$0.9 & 62.0$\pm$0.8 & 61.1$\pm$0.5 & 61.1$\pm$0.5 & 62.6$\pm$1.0 & 62.4$\pm$0.4 & 63.3$\pm$0.4 \\
        & $U_{\text{EigV}}$ & 66.3$\pm$0.7 & 67.9$\pm$0.5 & 69.0$\pm$0.6 & 69.5$\pm$0.9 & 67.9$\pm$0.4 & 68.8$\pm$0.2 & 69.7$\pm$1.3 & 70.0$\pm$0.7 & 70.9$\pm$0.5 \\
        & Variance (Tot) & 66.8$\pm$0.5 & 68.2$\pm$0.5 & 68.9$\pm$0.6 & 69.6$\pm$0.9 & 69.2$\pm$0.6 & 70.2$\pm$0.6 & 70.8$\pm$1.5 & 71.1$\pm$0.7 & 72.0$\pm$0.8 \\
        & Variance (AU) & 66.8$\pm$0.5 & 66.9$\pm$0.6 & 68.2$\pm$0.9 & --- & 69.2$\pm$0.6 & 70.3$\pm$0.6 & 70.7$\pm$1.6 & 71.3$\pm$0.8 & 71.6$\pm$0.9 \\
        & Variance (EU) & --- & 68.2$\pm$0.5 & 68.8$\pm$0.8 & 69.6$\pm$0.9 & --- & 68.0$\pm$0.6 & 70.7$\pm$1.6 & 71.3$\pm$0.8 & 71.6$\pm$0.9 \\
		\bottomrule
	\end{tabular}%
 	\caption{Calibration (AUROC) results of GPT-3.5 over $(n_{\text{p}}, n_{\text{s}})$ pairs.}
	\label{Table:Data1_overall}%
\end{table*}%

\subsection{Prompt Sensitivity Ratio}
We lastly introduce a simple metric that can be used to measure how much of a model's total uncertainty can be attributed to epistemic uncertainty:
\begin{equation}
\fontsize{7pt}{8pt}\selectfont
\rho_{\text{u}} = \frac{U_{e} + \epsilon}{U_{t} + 2\epsilon},
\label{eq:uncertainty_ratio}
\end{equation}
where $\epsilon$ is included to ensure $\rho_{\text{u}}$ is defined at $U_{t} = 0.$ In practice, we use $\epsilon = \text{1e-4}$. This ratio quantifies the proportion of total uncertainty attributed to epistemic uncertainty and thus how prompt-sensitive a model is at a given semantic concept. In Section \ref{case-study} we show how this ratio can be used to diagnose model overfitting.

\section{Experiments}
\subsection{Perturbation Method}
Let \(Q(Y \mid x, \theta)\) be the response distribution of the language model at an input $x$ and parameterization $\theta$. When using paraphrasing perturbations, $\theta$ refers to some paraphrase of the input $x.$ We consider a setting where we are allowed $m$ total samples that are distributed equally across $n_{\text{p}}$ perturbations \( \{\theta_1, \theta_2, \dots, \theta_{n_{\text{p}}}\} \). At each perturbation $\theta_i$, we take  $n_{\text{s}}$ responses \( \{y_{i1}, y_{i2}, \dots, y_{in_{\text{s}}}\} \sim Q(Y \mid x, \theta_i) \), such that \( m = n_{\text{p}} \cdot n_{\text{s}}\) (see Figure \ref{fig:2}). In Section \ref{perturbation-method-comparison} we compare different perturbation methods. For these experiments, $m = 6$ with $n_{\text{p}} = 6$ and $n_{\text{s}} = 1$. The baseline for these experiments, as well as the fixed-temperature study, reflects no-perturbation sampling at $m=6$, i.e., with the parameters $n_{\text{p}} = 1$ and $n_{\text{s}} = 6.$ In Section \ref{calibration-results} we attempt to answer the question, \textit{Given a fixed number of samples from the LLM, how should they optimally be distributed over perturbations?} To do this, we measure calibration at different factorizations of $m$ over $(n_{\text{p}}, n_{\text{s}})$. For $m = 6,$ we test the settings \(\{(1, 6), (2, 3), (3, 2), (6, 1)\}.\) For $m = 12,$ we test the settings \(\{(1, 12), (2, 6), (3, 4), (4, 3), (6, 2)\}.\)

\subsection{Experimental Setup}
\label{experimental_setup}
\textbf{Evaluation metrics.} We evaluate uncertainty calibration using AUROC when predicting the accuracy of a generation from its uncertainty as described in Section \ref{uq}. Following \citet{lin2023generating}, we evaluate the accuracy of generations using GPT-3.5 Turbo to grade. The model is provided with the question, correct answer, and generated answer, and asked to determine if it is correct.


\textbf{Datasets.} We use two question-answering datasets: TriviaQA \cite{joshi2017triviaqa} and Natural Questions (NQ) \cite{kwiatkowski2019natural}. 

\textbf{Baselines.} We benchmark our metric against several baseline metrics. Entropy is used as a common measure of uncertainty (Equation \ref{entropy}). Additionally, we include lexical similarity (LexiSim), semantic entropy (SE) from \citet{kuhn2023semantic}, and the sum of eigenvalues ($U_{\text{EigV}}$) metric based on the entailment affinity graph proposed by \citet{lin2023generating}.

\textbf{Implementation details.} For our experiments, we select the first 1,000 question-answer pairs from the validation split of each dataset. 
The number of perturbations ($n_{p}$) and the number of samples ($n_{s}$) were experimented with in different pairs. 
We run each experiment 5 times, and report the means and standard deviations of evaluation metrics. We utilize three LLMs: GPT-3.5 (gpt-3.5-turbo-0125) \cite{ouyang2022training}, Llama 2-Base (7B), and Llama 2-Chat (7B) \cite{touvron2023llama}. For GPT-3.5, we sample with temperature 1. For Llama models, we sample with temperature 0.6. The experiments were conducted on eight RTX A6000 GPUs.

\subsection{Perturbation Method Comparison}
\label{perturbation-method-comparison}
Previous work has proposed the use of temperature sampling to recover calibration in miscalibrated language models \cite{desai-durrett-2020-calibration, gao2024spuq, kadavath2022language, ling2024uncertainty}. However, the effects of temperature scaling on calibration can be misleading. By sampling at a high temperature, one can ``smooth" the predictive distribution, making the model systematically under-confident. For example, a model that always predicts 50\% probability on a binary, balanced dataset will be well-calibrated but have equal-to-random accuracy.

Previous work has evaluated uncertainty calibration in isolation. We additionally study the relationship between calibration and accuracy for different sampling methods. We consider several meaning-preserving perturbation methods: Paraphrasing (different phrasings of the same question), Dummy Tokens (adding irrelevant tokens to the question), and System Messages (using different system instructions).


We study the effects of temperature on calibration in two ways. First, following \citet{gao2024spuq}, we test perturbation by Random Temperature (sampling temperature randomly over the uniform distribution [0,1]). Second, also following \citet{gao2024spuq}, we test calibration performance across a range of fixed temperature settings, illustrating the trade-off between accuracy and calibration for temperature reshaping.

\begin{figure}[!htbp]
    \centering
    \begin{subfigure}[b]{\linewidth}
        \centering
        \includegraphics[width=.95\linewidth]{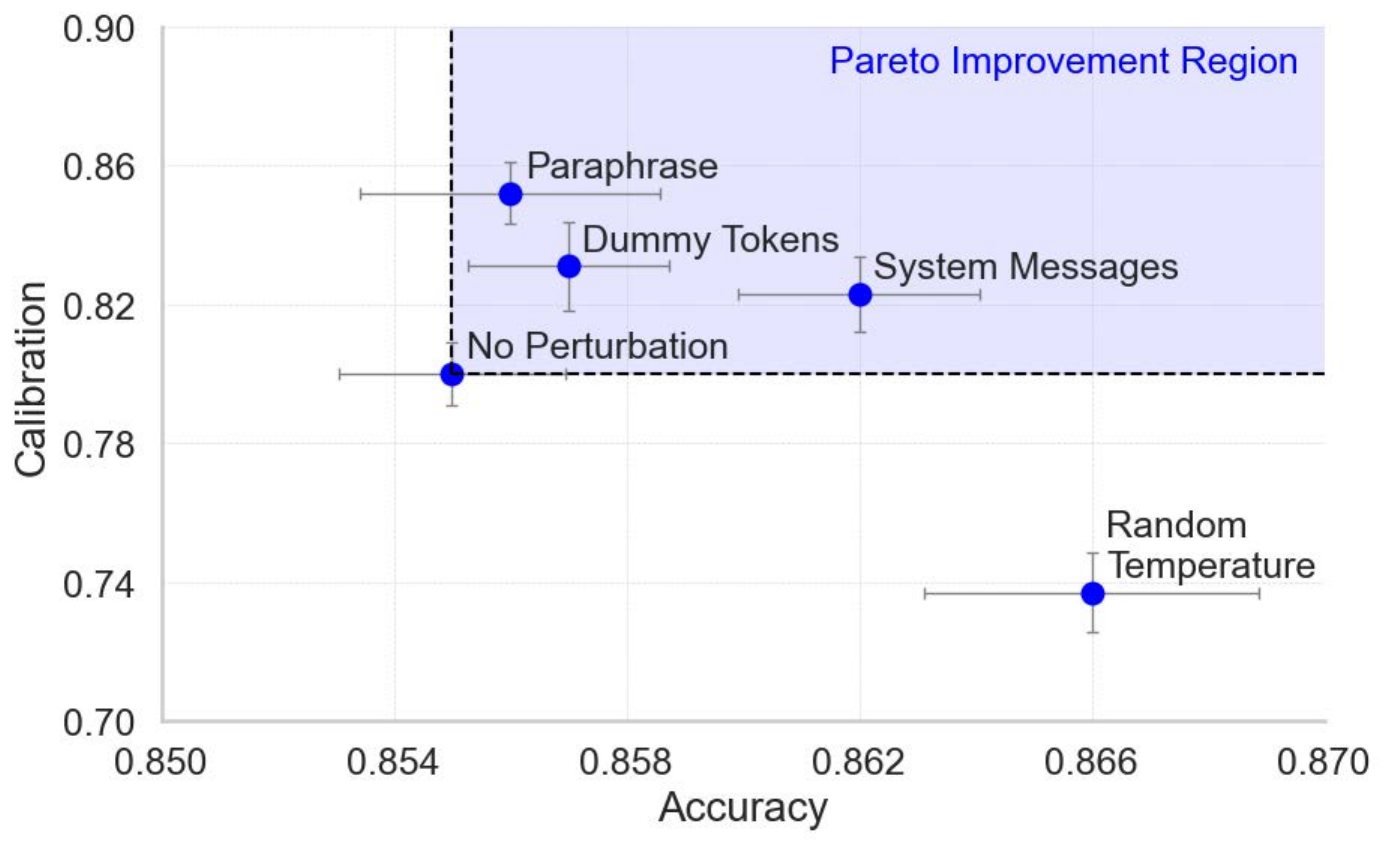}
        \caption{Calibration and accuracy (Temperature = 1).}
        \label{fig:first}
    \end{subfigure}
    
    \vspace{1em} 
    
    \begin{subfigure}[b]{\linewidth}
        \centering
        \includegraphics[width=.95\linewidth]{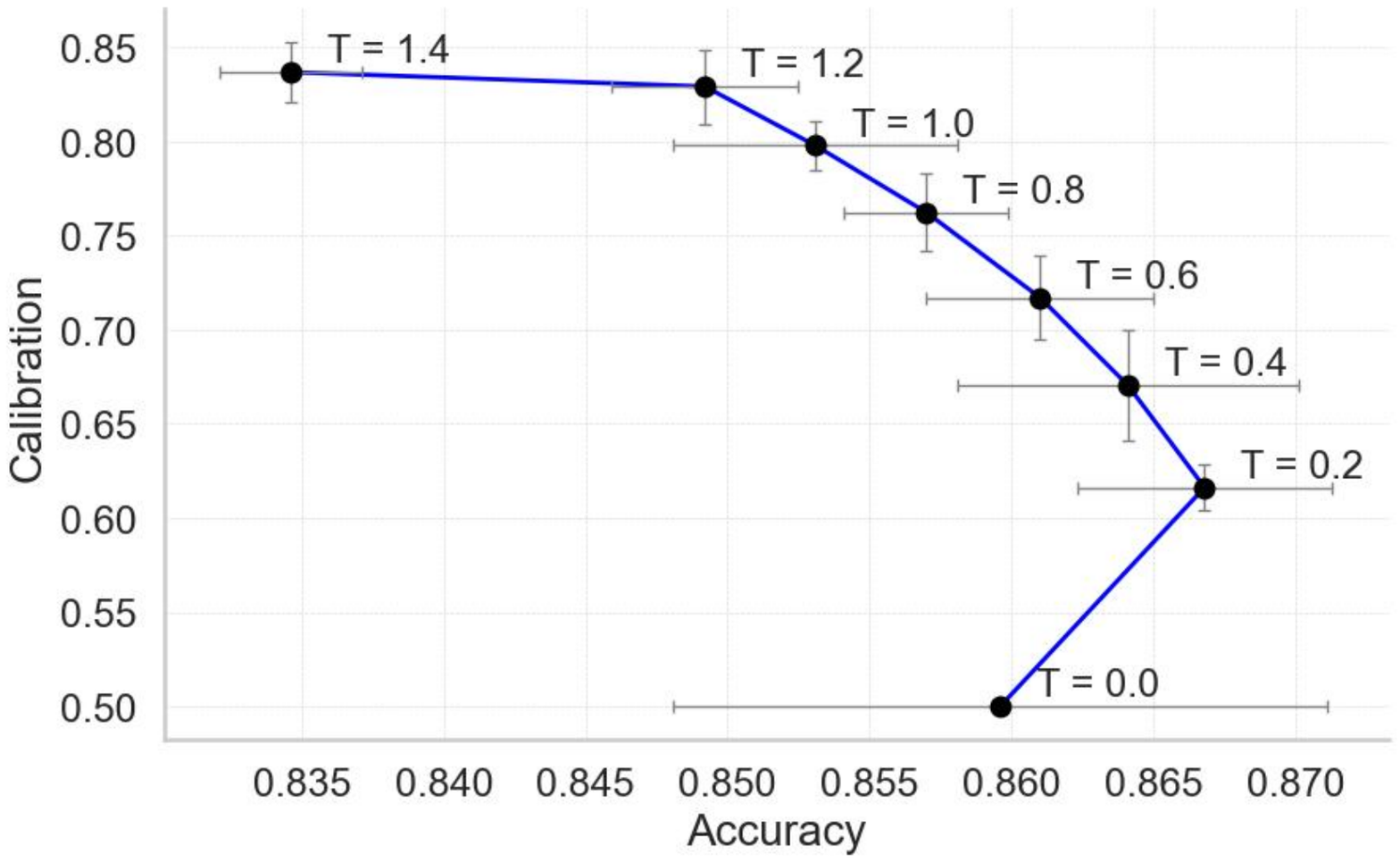}
        \caption{Calibration and accuracy at fixed sampling temperatures.}
        \label{fig:second}
    \end{subfigure}
    
    \caption{Comparison of calibration and accuracy at different temperatures.}
    \label{fig:combined}
\end{figure}

Figures \ref{fig:first} and \ref{fig:second} show the results of these experiments over 1,000 questions with GPT-3.5. For all perturbation methods in Figure \ref{fig:first}, for each question, we perform six perturbations, and sample from each perturbation once. Perturbations are compared to a no-perturbation baseline, in which we sample from a single, randomly chosen paraphrase 6 times for each question.

In the fixed temperature sampling results in Figure \ref{fig:second} we also sample from a single, randomly chosen paraphrase 6 times for each question at the indicated temperature setting. Figures \ref{fig:first} and \ref{fig:second} show the mean calibration (AUROC) and accuracy results over five experiments, with error bars representing the standard deviations.

\begin{table*}
    \centering
    \setlength{\tabcolsep}{1.0mm} 
    \fontsize{9}{11}\selectfont
    \begin{tabular}{lccccccccc}
        \toprule
        & \multicolumn{4}{c}{Calibration (AUROC)} & \multicolumn{5}{c}{$\rho_{\text{u}}$} \\
        \cmidrule(lr){2-5} \cmidrule(lr){6-10}
        & \multicolumn{2}{c}{TriviaQA} & \multicolumn{2}{c}{NQ} & \multicolumn{2}{c}{TriviaQA} & \multicolumn{2}{c}{NQ} \\
        & Base & Chat & Base & Chat & Base & Chat & Base & Chat & Baseline\\
        \midrule
        (1 , 12)  & $87.5 \pm 0.3$ & $79.2 \pm 0.5$ & $77.8 \pm 0.1$ & $68.1 \pm 0.6$ & --- & --- & --- & --- & --- \\
        (2 , 6) & $87.8 \pm 0.6$ & $81.8 \pm 0.5$ & $78.3 \pm 0.2$ & $71.1 \pm 0.5$ & $34.0 \pm 0.1$ & $38.0 \pm 0.2$ & $25.5 \pm 0.5$ & $32.1 \pm 0.5$ & $16.7$\\
        (3 , 4) & $88.1 \pm 0.2$ & $82.8 \pm 0.4$ & $78.1 \pm 0.6$ & $71.6 \pm 0.5$ & $38.0 \pm 0.4$ & $40.7 \pm 0.4$ & $32.0 \pm 0.2$ & $36.5 \pm 0.2$ & $25.0$ \\
        (4 , 3) & $88.1 \pm 0.5$ & $82.5 \pm 0.5$ & $77.9 \pm 0.5$ & $72.9 \pm 0.4$ & $44.2 \pm 0.2$ & $46.3 \pm 0.3$ & $40.3 \pm 0.2$ & $43.9 \pm 0.3$ & $33.3$ \\
        (6 , 2) & $88.1 \pm 0.7$ & $83.6 \pm 0.5$ & $77.9 \pm 0.4$ & $73.1 \pm 0.7$ & $52.8 \pm 0.2$ & $53.5 \pm 0.2$ & $53.4 \pm 0.3$ & $54.6 \pm 0.2$ & $50.0$ \\
        \bottomrule
    \end{tabular}
    \caption{Llama 2-Chat and Llama 2-Base comparison over $(n_{\text{p}}, n_{\text{s}})$ pairs.}
    \label{Table:table-2}
\end{table*}

Our overfitting framework for model miscalibration motivates recovering calibration by sampling across \textit{different} distributions, representing the same semantic concept. Changes in temperature sampling, however, sample from the same distribution, merely reshaped. We show that improvements in calibration due to temperature perturbation coincide with a decrease in accuracy. By contrast, semantic-invariant perturbation methods yield Pareto improvements in calibration (i.e., without trading accuracy). Further, paraphrasing perturbation, the most aggressive of the semantic-invariant perturbations, shows the greatest improvement in calibration. Our results reinforce that perturbation sampling in LLMs should use semantic-invariant perturbations, and suggest that paraphrasings are the optimal way to do this.

\subsection{Calibration Results}
\label{calibration-results}
The previous section demonstrates that perturbation by paraphrasing yields the greatest improvement in calibration, and does so without trading accuracy. Here, we measure how different degrees of paraphrasing perturbation affect model calibration. We also compare the calibration of the embedding variance metrics proposed in Section \ref{embedding-variance} to several baseline metrics: predictive entropy (Equation \ref{entropy}), lexical similarity (following \citet{lin2023generating}), semantic entropy from \citet{kuhn2023semantic} (Equation \ref{semantic-entropy}), and $U_{\text{EigV}}$ from \citet{lin2023generating}. Results in Table \ref{Table:Data1_overall} show that our embedding variance metric is competitive with the $U_{\text{EigV}}$ and surpasses all others.


As we increase the number of perturbations, we observe a shift in calibration attribution: epistemic uncertainty contributes more to calibration while aleatoric uncertainty contributes less. Sampling more broadly across the semantic concept space (through diverse paraphrases) improves calibration across all metrics. Notably, a sample from a new paraphrase provides more marginal information than an additional sample of the same paraphrase. This aligns with our understanding of token-level overfitting: The model's calibration suffers when repeatedly sampling from the same token representation, but improves when sampling across varied paraphrases that better capture the underlying distribution over the input's meaning.

Table 2 presents results from evaluation on Llama 2-Base and Llama 2-Chat. It includes calibration results based on total embedding variance, and the uncertainty ratio \( \rho_{\text{u}} \) from Equation \ref{eq:uncertainty_ratio} over choices of \( (n_{\text{p}}, n_{\text{s}}) \) for \( m = 12.\)

\maketitle
\subsection{Case Study: Base vs. Chat}
\label{case-study}
A common approach to fine-tuning a pre-trained LLM is reinforcement learning from human feedback (RLHF) \cite{christiano2017deep}. However, previous work has shown RLHF policies can cause models to become miscalibrated by ``collapsing" their output distributions \cite{openai2024gpt4, kadavath2022language}. In this section, we compare the calibrations of a pre-trained base model Llama 2-Base and 
the RLHF-fine-tuned model Llama 2-Chat \cite{touvron2023llama}.

When decomposing uncertainty using variance, we observe that the prompt sensitivity metric $\rho_{\text{u}}$ (the ratio of inter-sample variance $U_{\text{a}}$ to total variance $U_{\text{t}}$) has a natural relationship with the number of samples $n_{\text{s}}$. A perfectly generalizing model will sample all answers from the same underlying distribution, regardless of input perturbations. For such a model, $\rho_{\text{u}}$ will approach $\frac{1}{n_{\text{s}}}$ as $n_{\text{s}}$ grows. This provides us with a theoretical baseline: any positive deviation from $\frac{1}{n_{\text{s}}}$ indicates a departure from perfect generalization, making $\frac{1}{n_{\text{s}}}$ a prompt-insensitive baseline for $\rho_{\text{u}}$.


We note two trends in Table \ref{Table:table-2}: (1) Perturbation does not substantially improve calibration in the base model, indicating that the base model is well-calibrated at individual paraphrases. (2) The ratio $\rho_{\text{u}}$ is consistently larger for the chat model than the base model, indicating that the chat model is more prompt-sensitive than the base model. These results provide evidence that large pre-trained models like Llama 2-Base generalize well from token representations to question concepts, but post-training may induce post-hoc overfitting.

\section{Conclusion}
Our work considers the hypothesis that miscalibration in LLMs stems from prompt sensitivity, and demonstrates that sampling across semantic concept space using paraphrases improves calibration on QA tasks. We introduce a new uncertainty metric to quantify different components of uncertainty under our perturbation framework, and use these to measure model prompt sensitivity. Supporting previous work, we find models post-trained with RLHF are particularly prompt-sensitive. We hope our work provides a way to improve uncertainty calibration in black-box LLMs, facilitating more trustworthy AI systems, and sheds light on an interesting limitation of post-trained LLMs. Future work may explore whether prompt-sensitivity indicates a deeper structural shift in how these models encode and retrieve knowledge or a superficial artifact of alignment-driven constraints.

\section*{Acknowledgements}
The authors would like to acknowledge funding support from the following sources: NIH OT2OD032581, NIH OTA-21-008, NSF 2333703, NSF 2303038.




\bibliography{LaTeX/aaai25}

\begin{thebibliography}{26}
\providecommand{\natexlab}[1]{#1}

\bibitem[{Bubeck et~al.(2023)Bubeck, Chandrasekaran, Eldan, Gehrke, Horvitz, Kamar, Lee, Lee, Li, Lundberg et~al.}]{bubeck2023sparks}
Bubeck, S.; Chandrasekaran, V.; Eldan, R.; Gehrke, J.; Horvitz, E.; Kamar, E.; Lee, P.; Lee, Y.~T.; Li, Y.; Lundberg, S.; et~al. 2023.
\newblock Sparks of artificial general intelligence: Early experiments with gpt-4.
\newblock \emph{arXiv preprint arXiv:2303.12712}.

\bibitem[{Chen et~al.(2024)Chen, Chi, Wang, and Zhou}]{chen2024premise}
Chen, X.; Chi, R.~A.; Wang, X.; and Zhou, D. 2024.
\newblock Premise Order Matters in Reasoning with Large Language Models.
\newblock \emph{arXiv preprint arXiv:2402.08939}.

\bibitem[{Christiano et~al.(2017)Christiano, Leike, Brown, Martic, Legg, and Amodei}]{christiano2017deep}
Christiano, P.~F.; Leike, J.; Brown, T.; Martic, M.; Legg, S.; and Amodei, D. 2017.
\newblock Deep reinforcement learning from human preferences.
\newblock \emph{Advances in neural information processing systems}, 30.

\bibitem[{Depeweg et~al.(2018)Depeweg, Hernández-Lobato, Doshi-Velez, and Udluft}]{depeweg2018decompositionuncertaintybayesiandeep}
Depeweg, S.; Hernández-Lobato, J.~M.; Doshi-Velez, F.; and Udluft, S. 2018.
\newblock Decomposition of Uncertainty in Bayesian Deep Learning for Efficient and Risk-sensitive Learning.
\newblock arXiv:1710.07283.

\bibitem[{Desai and Durrett(2020)}]{desai-durrett-2020-calibration}
Desai, S.; and Durrett, G. 2020.
\newblock Calibration of Pre-trained Transformers.
\newblock In \emph{Proceedings of the 2020 Conference on Empirical Methods in Natural Language Processing (EMNLP)}, 295--302.

\bibitem[{Duan et~al.(2024)Duan, Cheng, Wang, Zavalny, Wang, Xu, Kailkhura, and Xu}]{duan2024shiftingattentionrelevancepredictive}
Duan, J.; Cheng, H.; Wang, S.; Zavalny, A.; Wang, C.; Xu, R.; Kailkhura, B.; and Xu, K. 2024.
\newblock Shifting Attention to Relevance: Towards the Predictive Uncertainty Quantification of Free-Form Large Language Models.
\newblock arXiv:2307.01379.

\bibitem[{Gao et~al.(2024)Gao, Zhang, Mouatadid, and Das}]{gao2024spuq}
Gao, X.; Zhang, J.; Mouatadid, L.; and Das, K. 2024.
\newblock SPUQ: Perturbation-Based Uncertainty Quantification for Large Language Models.
\newblock In \emph{Proceedings of the 18th Conference of the European Chapter of the Association for Computational Linguistics (Volume 1: Long Papers)}, 2336--2346.

\bibitem[{Hawkins(2021)}]{Hawkins2021-ak}
Hawkins, J. 2021.
\newblock \emph{Ergodic dynamics}.
\newblock Springer.

\bibitem[{Hou et~al.(2023)Hou, Liu, Qian, Andreas, Chang, and Zhang}]{hou2023decomposing}
Hou, B.; Liu, Y.; Qian, K.; Andreas, J.; Chang, S.; and Zhang, Y. 2023.
\newblock Decomposing uncertainty for large language models through input clarification ensembling.
\newblock \emph{arXiv preprint arXiv:2311.08718}.

\bibitem[{Joshi et~al.(2017)Joshi, Choi, Weld, and Zettlemoyer}]{joshi2017triviaqa}
Joshi, M.; Choi, E.; Weld, D.~S.; and Zettlemoyer, L. 2017.
\newblock TriviaQA: A Large Scale Distantly Supervised Challenge Dataset for Reading Comprehension.
\newblock In \emph{Proceedings of the 55th Annual Meeting of the Association for Computational Linguistics (Volume 1: Long Papers)}, 1601--1611.

\bibitem[{Kadavath et~al.(2022)Kadavath, Conerly, Askell, Henighan, Drain, Perez, Schiefer, Hatfield-Dodds, DasSarma, Tran-Johnson et~al.}]{kadavath2022language}
Kadavath, S.; Conerly, T.; Askell, A.; Henighan, T.; Drain, D.; Perez, E.; Schiefer, N.; Hatfield-Dodds, Z.; DasSarma, N.; Tran-Johnson, E.; et~al. 2022.
\newblock Language models (mostly) know what they know.
\newblock \emph{arXiv preprint arXiv:2207.05221}.

\bibitem[{Kaplan et~al.(2020)Kaplan, McCandlish, Henighan, Brown, Chess, Child, Gray, Radford, Wu, and Amodei}]{kaplan2020scalinglawsneurallanguage}
Kaplan, J.; McCandlish, S.; Henighan, T.; Brown, T.~B.; Chess, B.; Child, R.; Gray, S.; Radford, A.; Wu, J.; and Amodei, D. 2020.
\newblock Scaling Laws for Neural Language Models.
\newblock arXiv:2001.08361.

\bibitem[{Kuhn, Gal, and Farquhar(2022)}]{kuhn2023semantic}
Kuhn, L.; Gal, Y.; and Farquhar, S. 2022.
\newblock Semantic Uncertainty: Linguistic Invariances for Uncertainty Estimation in Natural Language Generation.
\newblock In \emph{The Eleventh International Conference on Learning Representations}.

\bibitem[{Kwiatkowski et~al.(2019)Kwiatkowski, Palomaki, Redfield, Collins, Parikh, Alberti, Epstein, Polosukhin, Devlin, Lee et~al.}]{kwiatkowski2019natural}
Kwiatkowski, T.; Palomaki, J.; Redfield, O.; Collins, M.; Parikh, A.; Alberti, C.; Epstein, D.; Polosukhin, I.; Devlin, J.; Lee, K.; et~al. 2019.
\newblock Natural questions: a benchmark for question answering research.
\newblock \emph{Transactions of the Association for Computational Linguistics}, 7: 453--466.

\bibitem[{Lin, Trivedi, and Sun(2023)}]{lin2023generating}
Lin, Z.; Trivedi, S.; and Sun, J. 2023.
\newblock Generating with confidence: Uncertainty quantification for black-box large language models.
\newblock \emph{arXiv preprint arXiv:2305.19187}.

\bibitem[{Ling et~al.(2024)Ling, Zhao, Zhang, Cheng, Liu, Sun, Oishi, Osaki, Matsuda, Ji, Bai, Zhao, and Chen}]{ling2024uncertainty}
Ling, C.; Zhao, X.; Zhang, X.; Cheng, W.; Liu, Y.; Sun, Y.; Oishi, M.; Osaki, T.; Matsuda, K.; Ji, J.; Bai, G.; Zhao, L.; and Chen, H. 2024.
\newblock Uncertainty Quantification for In-Context Learning of Large Language Models.
\newblock \emph{arXiv preprint arXiv:2402.10189}.

\bibitem[{Lu et~al.(2024)Lu, Bigoulaeva, Sachdeva, Madabushi, and Gurevych}]{lu2024emergentabilitieslargelanguage}
Lu, S.; Bigoulaeva, I.; Sachdeva, R.; Madabushi, H.~T.; and Gurevych, I. 2024.
\newblock Are Emergent Abilities in Large Language Models just In-Context Learning?
\newblock arXiv:2309.01809.

\bibitem[{Lu et~al.(2022)Lu, Bartolo, Moore, Riedel, and Stenetorp}]{Lu2022Fantastically}
Lu, Y.; Bartolo, M.; Moore, A.; Riedel, S.; and Stenetorp, P. 2022.
\newblock Fantastically Ordered Prompts and Where to Find Them: Overcoming Few-Shot Prompt Order Sensitivity.
\newblock In \emph{Proceedings of the 60th Annual Meeting of the Association for Computational Linguistics (Volume 1: Long Papers)}, 8086--8098.

\bibitem[{McCoy et~al.(2023)McCoy, Yao, Friedman, Hardy, and Griffiths}]{mccoy2023embersautoregressionunderstandinglarge}
McCoy, R.~T.; Yao, S.; Friedman, D.; Hardy, M.; and Griffiths, T.~L. 2023.
\newblock Embers of Autoregression: Understanding Large Language Models Through the Problem They are Trained to Solve.
\newblock arXiv:2309.13638.

\bibitem[{Ng, Jordan, and Weiss(2001)}]{ng2001spectral}
Ng, A.; Jordan, M.; and Weiss, Y. 2001.
\newblock On spectral clustering: Analysis and an algorithm.
\newblock \emph{Advances in neural information processing systems}, 14.

\bibitem[{OpenAI(2023)}]{openai2024gpt4}
OpenAI. 2023.
\newblock GPT-4 Technical Report.
\newblock \emph{arXiv preprint arXiv:2303.08774}.

\bibitem[{Ouyang et~al.(2022)Ouyang, Wu, Jiang, Almeida, Wainwright, Mishkin, Zhang, Agarwal, Slama, Ray et~al.}]{ouyang2022training}
Ouyang, L.; Wu, J.; Jiang, X.; Almeida, D.; Wainwright, C.; Mishkin, P.; Zhang, C.; Agarwal, S.; Slama, K.; Ray, A.; et~al. 2022.
\newblock Training language models to follow instructions with human feedback.
\newblock \emph{Advances in neural information processing systems}, 35: 27730--27744.

\bibitem[{Power et~al.(2022)Power, Burda, Edwards, Babuschkin, and Misra}]{power2022grokking}
Power, A.; Burda, Y.; Edwards, H.; Babuschkin, I.; and Misra, V. 2022.
\newblock Grokking: Generalization beyond overfitting on small algorithmic datasets.
\newblock \emph{arXiv preprint arXiv:2201.02177}.

\bibitem[{Sclar et~al.(2023)Sclar, Choi, Tsvetkov, and Suhr}]{Sclar2023Sensitivity}
Sclar, M.; Choi, Y.; Tsvetkov, Y.; and Suhr, A. 2023.
\newblock Quantifying Language Models' Sensitivity to Spurious Features in Prompt Design or: How I learned to start worrying about prompt formatting.
\newblock In \emph{The Twelfth International Conference on Learning Representations}.

\bibitem[{Shi et~al.(2023)Shi, Chen, Misra, Scales, Dohan, Chi, Sch{\"a}rli, and Zhou}]{shi2023large}
Shi, F.; Chen, X.; Misra, K.; Scales, N.; Dohan, D.; Chi, E.~H.; Sch{\"a}rli, N.; and Zhou, D. 2023.
\newblock Large language models can be easily distracted by irrelevant context.
\newblock In \emph{International Conference on Machine Learning}, 31210--31227. PMLR.

\bibitem[{Touvron et~al.(2023)Touvron, Martin, Stone, Albert, Almahairi, Babaei, Bashlykov, Batra, Bhargava, Bhosale et~al.}]{touvron2023llama}
Touvron, H.; Martin, L.; Stone, K.; Albert, P.; Almahairi, A.; Babaei, Y.; Bashlykov, N.; Batra, S.; Bhargava, P.; Bhosale, S.; et~al. 2023.
\newblock Llama 2: Open foundation and fine-tuned chat models.
\newblock \emph{arXiv preprint arXiv:2307.09288}.

\end{thebibliography}

\end{document}